# Convexifying the Bethe Free Energy

Ofer Meshi    Ariel Jaimovich    Amir Globerson    Nir Friedman

School of Computer Science and Engineering
Hebrew University, Jerusalem, Israel 91904
{meshi,arielj,gamir,nir}@cs.huji.ac.il

**Abstract**

The introduction of loopy belief propagation (LBP) revitalized the application of graphical models in many domains. Many recent works present improvements on the basic LBP algorithm in an attempt to overcome convergence and local optima problems. Notable among these are convexified free energy approximations that lead to inference procedures with provable convergence and quality properties. However, empirically LBP still outperforms most of its convex variants in a variety of settings, as we also demonstrate here. Motivated by this fact we seek convexified free energies that directly approximate the Bethe free energy. We show that the proposed approximations compare favorably with state-of-the art convex free energy approximations.

## 1 Introduction

Computing likelihoods and marginal probabilities is a critical subtask in applications of graphical models. As computing the exact answers is often infeasible, there is a growing need for approximate inference methods. An important class of approximations are *variational methods* [7] that pose inference in terms of minimizing the free energy functional. In the last decade, loopy belief propagation (LBP), a simple local message passing procedure, proved to be empirically successful and was used in a variety of applications [10]. The seminal work of Yedidia *et al.* [20] merged these lines of work by formulating loopy belief propagation in terms of optimizing the Bethe free energy, an approximate free energy functional.

LBP suffers from two inherent problems: it fails to converge in some cases, and may converge to local optima due to the non-convexity of the Bethe free energy. Several approaches have been introduced to fix the non-convergence issue, so that LBP provably converges to a local optimum of the Bethe free energy [17, 22]. However, this still leaves the problem of local optima, and therefore the dependence of the solution on initial conditions. To alleviate this problem, several works [2, 6, 13, 15] construct convex free energy approximations, for which there is a single global optimum. Convexity also paved the way for the introduction of provably convergent message-passing algorithms for calculating likelihood and marginal probabilities [3, 4]. Moreover, some of these approximations provide upper bounds on the partition function [2, 13].

Despite their algorithmic elegance and convergence properties, convex variants often do not provide better empirical results than LBP. While this observation is shared by many practitioners, it does not have firm theoretical justification. Motivated by this observation, our goal in this work is to construct approximations that are both convex *and* directly approximate the Bethe free energy. We show how to approximate the Bethe in L2 norm and how to find the best upper bound on it for a given random field. We then illustrate the utility of our proposed approximations by comparing them to previously suggested ones across a variety of models and parameterizations.

## 2 Free Energy Approximations

Probabilistic graphical models provide a succinct language to specify complex joint probabilities over many variables. This is done by factorizing the distribution into a product over local potentials. Let $\mathbf{x} \in \mathcal{X}^n$ denote a vector of $n$ discrete random variables. Here we focus on Markov Random Fields where the joint probability is given by:

$$p(\mathbf{x};\boldsymbol{\theta}) = \frac{1}{Z(\boldsymbol{\theta})} \exp\left\{\sum_{\alpha} \theta_\alpha(x_\alpha) + \sum_{i} \theta_i(x_i)\right\} \quad (1)$$

where $\alpha$ correspond to subsets of parameters (or factors), and $Z(\boldsymbol{\theta})$ is the *partition function* that serves to normalize the distribution. We denote *marginal distributions* over variables and factors by $\mu_i(x_i)$ and $\mu_\alpha(x_\alpha)$, respectively, and the vector of all marginals by $\boldsymbol{\mu}$.

Given such a model, we are interested in computing the marginal probabilities $\boldsymbol{\mu}$, as well as the partition function

$Z(\boldsymbol{\theta})$, which is required for calculating the likelihood of evidence, especially in the context of parameter estimation. Finding exact answers for these tasks is theoretically and practically hard and thus many works often resort to approximate inference.

A class of popular approximate inference approaches are the variational methods that rely on the following exact formulation of $\log Z(\boldsymbol{\theta})$ [14]:

$$\log Z(\boldsymbol{\theta}) = \max_{\boldsymbol{\mu} \in \mathcal{M}(G)} \left\{ \boldsymbol{\theta}^T \boldsymbol{\mu} + H(\boldsymbol{\mu}) \right\} \qquad (2)$$

The set $\mathcal{M}(G)$ is known as the *marginal polytope* associated with a graph $G$ [14]. A vector $\boldsymbol{\mu}$ is in $\mathcal{M}(G)$ if it corresponds to the marginals of *some* distribution $p(\mathbf{x})$:

$$\mathcal{M}(G) = \left\{ \boldsymbol{\mu} \Big| \exists p(\mathbf{x}) \text{ s.t. } \begin{array}{l} \mu_i(x_i) = p(x_i) \\ \mu_\alpha(x_\alpha) = p(x_\alpha) \end{array} \right\} \qquad (3)$$

$H(\boldsymbol{\mu})$ is defined as the entropy of the unique exponential distribution of the form in Eq. (1) consistent with marginals $\boldsymbol{\mu}$. Finally, the objective in Eq. (2) is the negative of the *free energy functional*, denoted $F[\boldsymbol{\mu}, \boldsymbol{\theta}]$.

The solution to the optimization problem in Eq. (2) is precisely the desired vector of marginals of the distribution $p(\mathbf{x}; \boldsymbol{\theta})$.

In itself, this observation is not sufficient to provide an efficient algorithm, since the maximization in Eq. (2) is as hard as the original inference task. Specifically, $\mathcal{M}(G)$ is difficult to characterize and the computation of $H(\boldsymbol{\mu})$ is also intractable, so both need to be approximated. First, one can relax the optimization problem to be over an outer bound on the marginal polytope. In particular, it is natural to require that the resulting *pseudo-marginals* obey some local normalization and marginalization constraints. These constraints define the *local polytope*

$$L(G) = \left\{ \boldsymbol{\mu} \geq 0 \Big| \begin{array}{l} \sum_{x_i} \mu_i(x_i) = 1 \\ \sum_{x_\alpha \setminus x_i} \mu_\alpha(x_\alpha) = \mu_i(x_i) \end{array} \right\} \qquad (4)$$

As for the entropy term, a family of entropy approximations with a long history in statistical physics is based on a weighted sum of local entropies $H_\mathbf{c}(\boldsymbol{\mu}) = \sum_r c_r H_r(\mu_r)$, where $r$ are subsets of variables (regions) and the coefficients $c_r$ are called *counting numbers* [21]. The approximate optimization problem then takes the form:

$$\log \tilde{Z}(\boldsymbol{\theta}) = \max_{\boldsymbol{\mu} \in L(G)} \left\{ \boldsymbol{\theta}^T \boldsymbol{\mu} + H_\mathbf{c}(\boldsymbol{\mu}) \right\} \qquad (5)$$

The entropy approximation is defined both by the choice of regions and by the choice of counting numbers. This poses two complementary challenges: defining the regions, and assigning counting numbers for these regions. Here we focus on the second problem, which arises for any choice of regions. For simplicity, we limit ourselves to a common choice of regions — over variables and factors, although the results to follow can be generalized to more elaborate region choices (*e.g.*, [16, 18]). In this case the approximate entropy takes the form:

$$H_\mathbf{c}(\boldsymbol{\mu}) = \sum_i c_i H_i(\mu_i) + \sum_\alpha c_\alpha H_\alpha(\mu_\alpha) \qquad (6)$$

where $c_i$ and $c_\alpha$ are the counting numbers for variables and factors, respectively.

Each set of counting numbers will result in a different approximation. The *Bethe* entropy approximation $H_\mathbf{b}(\boldsymbol{\mu})$ is defined by choosing $c_\alpha = 1$, $c_i = 1 - d_i$ (where $d_i = |\{\alpha : i \in \alpha\}|$) [21].

**Concave Entropy Approximations**

One shortcoming of the Bethe entropy is that it is not concave, and thus Eq. (5) may have local optima. It is possible to consider instead entropy approximations that are provably concave. Such approximations have been studied extensively in recent years, along with provable convergent algorithms for solving Eq. (5) in these cases. One of the first concave entropy approximations introduced was the tree-reweighting (TRW) method of Wainwright *et al.*[13]. The TRW entropy approximation is a convex combination of tree entropies and is concave. Furthermore, it is an upper bound on the true $H(\boldsymbol{\mu})$ so that the optimum of Eq. (5) yields an upper bound on $\log Z(\boldsymbol{\theta})$.

More recently, Heskes [5] derived a set of sufficient condition for $c_\alpha, c_i$ to yield a concave function. He showed that an entropy approximation $H_\mathbf{c}(\boldsymbol{\mu})$ is provably concave for $\boldsymbol{\mu} \in L(G)$ if there exist auxiliary counting numbers $c_{\alpha\alpha}, c_{ii}, c_{i\alpha} \geq 0$ such that

$$c_\alpha = c_{\alpha\alpha} + \sum_{i: i \in \alpha} c_{i\alpha} \quad \forall_\alpha \qquad (7)$$

$$c_i = c_{ii} - \sum_{\alpha: i \in \alpha} c_{i\alpha} \quad \forall_i \qquad (8)$$

## 3 Message Passing Algorithms

The optimization problem in Eq. (5) can be solved using generic optimization tools. However, message passing algorithms have proved especially useful for this task. Starting with the work of Yedidia *et al.*[20] many message passing algorithms have been proposed for optimizing variational approximations. Although not all these algorithms are provably convergent, if they do converge, it is to a fixed point of Eq. (5). Furthermore if the entropy $H_\mathbf{c}(\boldsymbol{\mu})$ is concave, this is the global optimum of Eq. (5).

Most existing algorithms make the assumption that $c_\alpha = 1$. Since in this work we want to explore a broader range of approximations, we derive message passing updates for the more general case (for any $c_\alpha \neq 0$). Our derivation, which follows closely the one of Yedidia *et al.*, results with the following update rules:

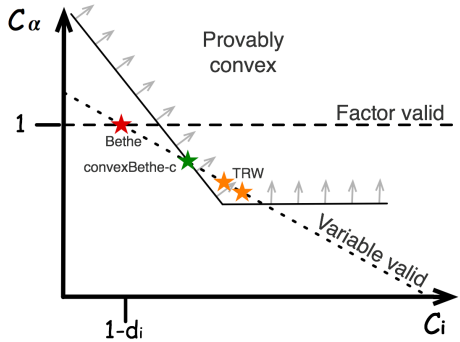

Figure 1: Illustration of the counting number space with different regions and point of interest labeled (see text).

$$m_{\alpha \to i}(x_i) = m^0_{\alpha \to i}(x_i)^{\frac{c_\alpha}{c_\alpha - q_i + 1}} n^0_{i \to \alpha}(x_i)^{\frac{q_i - c_\alpha}{c_\alpha - q_i + 1}} \quad (9)$$

$$n_{i \to \alpha}(x_i) = m^0_{\alpha \to i}(x_i)^{\frac{q_i - 1}{c_\alpha - q_i + 1}} n^0_{i \to \alpha}(x_i)^{\frac{1}{c_\alpha - q_i + 1}} \quad (10)$$

Where $q_i = \frac{1 - c_i}{d_i}$ and:

$$m^0_{\alpha \to i}(x_i) = \sum_{x_\alpha \setminus x_i} e^{\frac{1}{c_\alpha} \theta_\alpha(x_\alpha)} \prod_{\substack{j \in \alpha \\ j \neq i}} n_{j \to \alpha}(x_j)$$

$$n^0_{i \to \alpha}(x_i) = e^{\theta_i(x_i)} \prod_{\substack{\beta : i \in \beta \\ \beta \neq \alpha}} m_{\beta \to i}(x_i)$$

Note that by plugging in the Bethe counting numbers, where $c_\alpha = 1$ and $c_i = 1 - d_i$, this reduces back to the standard BP messages. Furthermore, if we set $c_\alpha = 1$ as in Yedidia *et al.* [21], then Eq. (9) and Eq. (10) reduce to the two-way algorithm defined there.

The above updates are not guaranteed to converge even if $H_\mathbf{c}(\boldsymbol{\mu})$ is concave. However, we have found that with dampening of messages it did converge for all the cases we studied.

## 4 Properties of Counting Number Space

Given a specific model, different choices of counting numbers lead to different entropy approximations. But what are *good* counting numbers and how can we find those for a specific model?

One desirable property of counting numbers is that they result in concave entropies, as discussed in Section 2. There are several rationales for choosing those. One is clearly that optimization is globally optimal. The other is that the true entropy $H(\boldsymbol{\mu})$ is itself concave [14].

Another approach to deriving good counting numbers is to restrict ourselves to $\mathbf{c}$ such that optimization with $H_\mathbf{c}(\boldsymbol{\mu})$ is exact for at least some values of $\boldsymbol{\theta}$. For example, suppose we have a model where $\theta_\alpha(x_\alpha) = 0$ for all $\alpha$ (i.e., a completely factorized model). How should we constrain $\mathbf{c}$ such that these models are optimized correctly? It is easy to see that if $\mathbf{c}$ satisfies:

$$c_i + \sum_{\alpha, i \in \alpha} c_\alpha = 1 \quad \forall i \quad (11)$$

then the corresponding $H_\mathbf{c}(\boldsymbol{\mu})$ will solve the factorized model exactly. We call $\mathbf{c}$ values that obey Eq. (11) *variable-valid*, as the variables are counted correctly in $H_\mathbf{c}(\boldsymbol{\mu})$ [21].

In a similar way we can define *factor-valid* approximations that satisfy:

$$c_\alpha = 1 \quad \forall \alpha \quad (12)$$

Intuitively, approximations that satisfy both Eq. (11) and Eq. (12) are *valid* [21] as they have the appealing property of not over- or under-counting variables and factors in the approximate entropy $H_\mathbf{c}(\boldsymbol{\mu})$. Furthermore, for tree structured distributions it has been shown that only valid counting numbers can yield exact results [11].

Figure 1 illustrates the structure of the above constraints in the space of counting numbers. Note that the Bethe approximation is the single choice of counting numbers that is both factor- and variable-valid. The TRW approximation is, by definition, always variable-valid, and any distribution over spanning trees results in a different value of $c_\alpha$. Finally, we note that for different model structures the counting number space looks different. For example, the Bethe approximation for tree structured distributions is convex.

To better understand the properties of different counting numbers, we perform an experiment where the counting number space is two dimensional and can be visualized. For this we use $5 \times 5$ toroidal grids in which each variable is connected to four neighboring variables by pairwise factors. The joint probability distribution of the model is given by: $p(x; \boldsymbol{\theta}) = \frac{1}{Z(\boldsymbol{\theta})} \exp \left\{ \sum_i \theta_i x_i + \sum_{(i,j) \in G} \theta_{i,j} x_i x_j \right\}$ with $x_i, x_j \in \{\pm 1\}$. The *field parameters* $\theta_i$ were drawn uniformly from the range $[-\omega_F, \omega_F]$, and the *interaction parameters* $\theta_{i,j}$ were drawn either from the range $[-\omega_I, \omega_I]$ or from $[0, \omega_I]$ to obtain *mixed* or *attractive* potentials respectively [2, 13]. This model structure has inherent symmetry as all factors are pairwise and each variable appears in exactly four factors. Hence, if we choose the counting numbers of the approximation based solely on the structure of the model, we get the same $c_i$ for all variables and the same $c_\alpha$ for all factors.

Figure 2 shows the performance of various approximations on two models. The first model is sampled in an easy regime with relatively weak interaction parameters while the second model is sampled from a more difficult regime with stronger interaction parameters.

We observe that most convex free energy approximations have large errors both in the estimate of the log-partition and in that of the marginal beliefs. When looking at subspaces that tend to empirically perform better than others, the convex subspace does not seem to generally give good approximations. However, we notice that variable-valid approximations stand out as the main region of relatively low error. In fact, we note that to the best of our knowledge all free energy approximations suggested in the literature obey this variable-valid constraint [2, 4, 13, 19, 21].

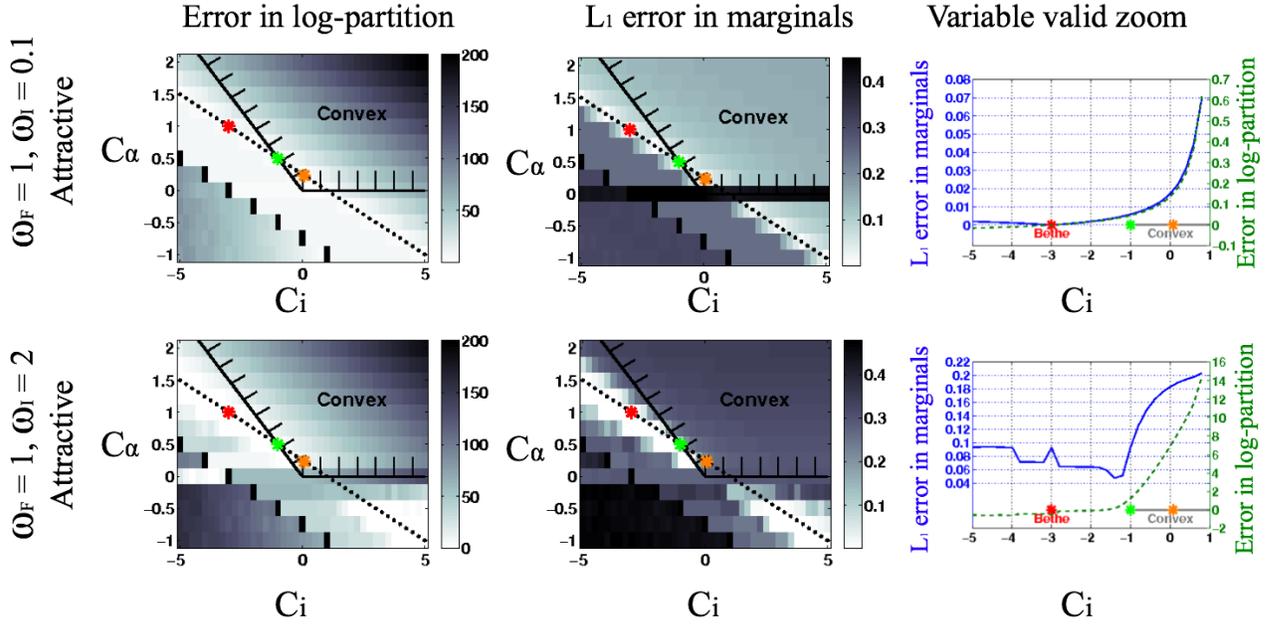

Figure 2: Quality of different approximations for two instances of the $5 \times 5$ toroidal grid from different parametric scenarios. In each matrix the $x$-axis denotes the counting number for nodes ($c_i$) and the $y$-axis denotes the counting number for edges of the grid ($c_\alpha$). Each pixel in the matrix is the result of running belief propagation with these counting numbers. The subspace of provably convex approximations is bound by solid lines, and the variable-valid subspace is marked by a dotted line. The left column shows the error in approximate log-partition function ($|\log Z(\boldsymbol{\theta}) - \log \tilde{Z}(\boldsymbol{\theta})|$), the middle column shows the average $L_1$ error in approximate marginals over factors and variables. The rightmost column shows in more detail the approximation quality in the variable-valid subspace. The colored stars show various approximations (see Figure 1 and text).

The rightmost column of Figure 2 shows performance of variable-valid approximations. We notice that for almost all models tested the approximation improves as the counting numbers get closer to the Bethe counting numbers. However, in most cases the Bethe approximation outperforms its convex counterparts. We obtained similar results for fully connected graphs and other non-pairwise models (not shown).

## 5 Approximating Bethe

The experiments in the previous section demonstrate that the Bethe free energy performs well across a variety of parameter settings. Since we would like to work with convex free energies, a key question is which of the convex free energies comes closest to the performance of Bethe.

In what follows, we describe several approaches to obtaining counting numbers that satisfy the above requirements. We divide these into *static* approximations that determine the counting numbers based only on the structure of the model, and *adaptive* approximations that also take the model parameters into account.

### 5.1 Static Approximations

Following the insights of the previous section, it is natural to try to combine the convexity constraints of Eq. (7) and Eq. (8) with the validity constraints defined by Eq. (11). Inside this subspace, a straightforward choice is to find the counting numbers that are closest in terms of Euclidean distance to the Bethe counting numbers. For clarity we denote by $\mathbf{b}$ the vector of Bethe counting numbers with $b_i = 1 - d_i$ and $b_\alpha = 1$. We define the *convexBethe-c* approximation as the solution to the optimization problem

$$\underset{\mathbf{c}}{\operatorname{argmin}} \|\mathbf{c} - \mathbf{b}\|^2 \qquad (13)$$

s.t. $c_i, c_\alpha$ satisfy Eq. (7,8,11).

This constrained optimization problem can be formulated as a quadratic program and solved using standard solvers. A similar approach was recently studied by Hazan and Shashua [4].

However, it is not clear that the L2 metric in counting number space is adequate for approximating Bethe. In principle we would like to approximate the Bethe entropy itself rather than its counting numbers. Since $H_\mathbf{b}(\boldsymbol{\mu})$ is a function of $\boldsymbol{\mu}$ we would like to find a function $H_\mathbf{c}(\boldsymbol{\mu})$ that is

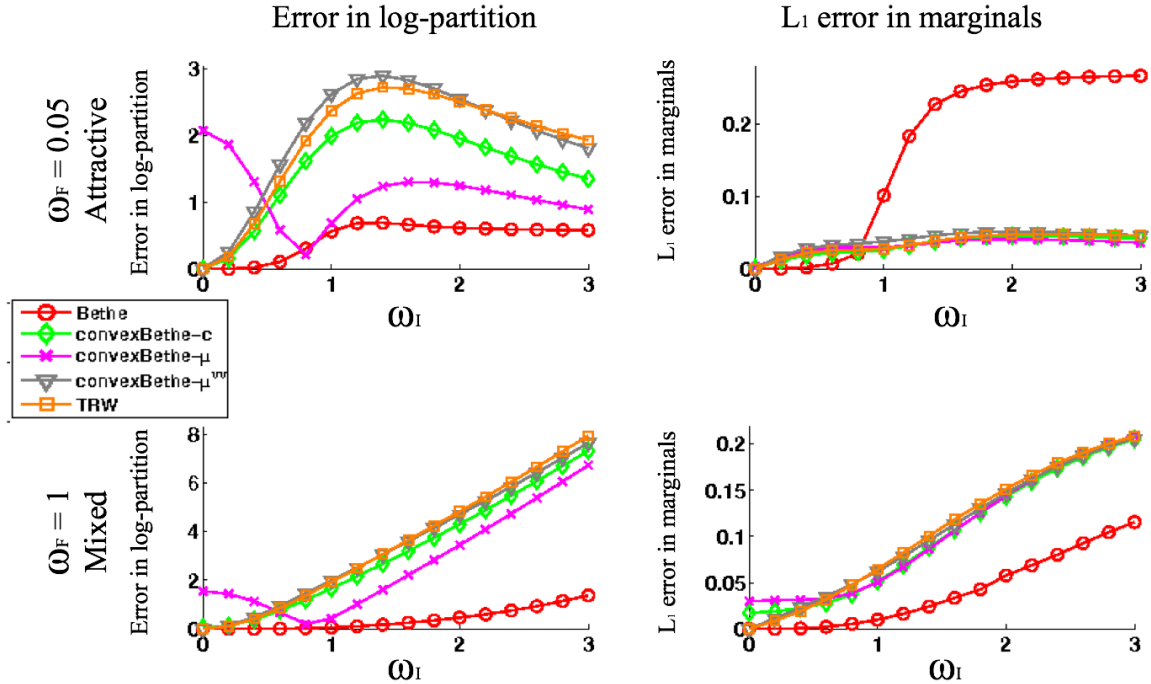

Figure 3: Comparison of estimation errors the partition function (left column) and the marginal probabilities (right column) for several approximation schemes. In the first row we take $\omega_F = 0.05$ and attractive potentials, and in the second row $\omega_F = 1$ and mixed potentials. In each graph the $x$-axis corresponds to the interaction meta-parameters $\omega_I$, and the $y$-axis shows the error in the log partition estimation (left column) and marginal estimation (right column). Each line describes the average errors over 20 $5 \times 5$ grid models sampled with the corresponding meta-parameters. The TRW approximation in this graph corresponds to the uniform distribution over four spanning trees.

closest to it when integrating over all $\boldsymbol{\mu}$ values. We can put this formally as:

$$\operatorname*{argmin}_{\mathbf{c}} \int_{L(G)} (H_{\mathbf{b}}(\boldsymbol{\mu}) - H_{\mathbf{c}}(\boldsymbol{\mu}))^2 \, d\boldsymbol{\mu} \qquad (14)$$

s.t. $c_i, c_\alpha$ satisfy Eq. (7,8).

We integrate over $L(G)$ since this is the optimization range in Eq. (5) and thus the relevant domain of approximation. Although this integration seems daunting, we can simplify the problem by noticing that $H_{\mathbf{b}}(\boldsymbol{\mu})$ and $H_{\mathbf{c}}(\boldsymbol{\mu})$ can be written as $\mathbf{b}^T \mathbf{H}_{\boldsymbol{\mu}}$ and $\mathbf{c}^T \mathbf{H}_{\boldsymbol{\mu}}$ respectively, where $\mathbf{H}_{\boldsymbol{\mu}}$ is the vector of local entropies. This results in the following quadratic optimization problem:

$$\operatorname*{argmin}_{\mathbf{c}} (\mathbf{b} - \mathbf{c})^T \mathbf{A} (\mathbf{b} - \mathbf{c}) \qquad (15)$$

s.t. $c_i, c_\alpha$ satisfy Eq. (7,8).

where

$$\mathbf{A} = \int_{L(G)} \mathbf{H}_{\boldsymbol{\mu}} \mathbf{H}_{\boldsymbol{\mu}}^T \, d\boldsymbol{\mu}$$

is the matrix of integration of all pairwise products of local entropy terms. Exact calculation of $\mathbf{A}$ is intractable and so we resort to MCMC methods[1], by performing a random walk inside $L(G)$. Starting with a random point inside $L(G)$ (i.e., a set of consistent marginals $\boldsymbol{\mu}$) we sample a legal direction, find the two boundaries along this direction, and then sample uniformly a new point from within the bounded interval. A straightforward argument shows that the stationary distribution of this walk is uniform within $L(G)$. To determine when the random walk is close to the stationary distribution, we apply a heuristic convergence test by running in parallel several chains from different random starting points and comparing their statistics [1]. Once we determine convergence, we then use samples from the different runs to estimate $\mathbf{A}$. Finally, we solve the optimization problem in Eq. (15) with and without enforcing the variable-valid constraints of Eq. (11), and term these approximations *convexBethe-$\mu^{\text{vv}}$* and *convexBethe-$\mu$*, respectively.

We evaluate the quality of these approximations for calculating the marginals and partition function in $5 \times 5$ non-toroidal grids with various parameterizations. We compare their performance with the Bethe approximation and the TRW approximation using a uniform distribution over four spanning trees (see [13]) [2]. Following Wainwright *et al.* [13], in each trial we set $\omega_F$ to a fixed value and gradually increase the interaction strength $\omega_I$. For each com-bination of $\omega_F$ and $\omega_I$ we sample 20 random models and measure the estimation errors of each approximation (see Figure 3).

---

[1] Volume computations over such polytopes are generally difficult, but in some special cases may be solved in closed form [8].

[2] We also used uniform TRW weights over all spanning trees and got similar results (not shown).

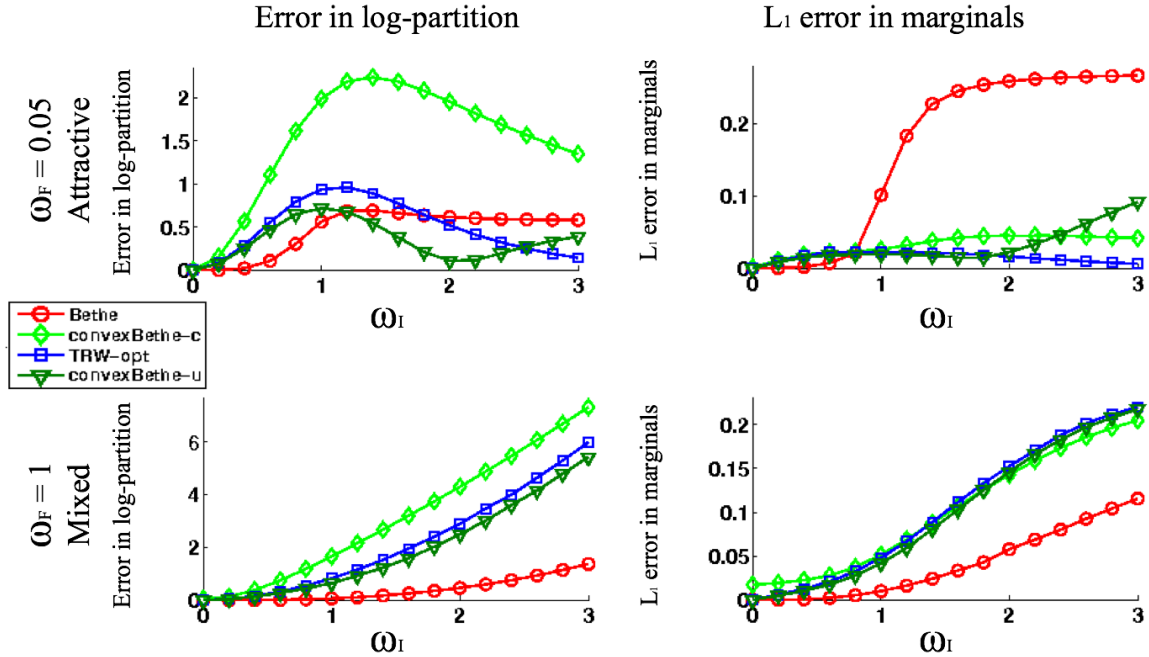

Figure 4: Comparison of estimation errors in partition function and marginals of the adaptive free energy approximations. The experimental setting is similar to that of Figure 3.

We first observe that the Bethe approximation outperforms its convex counterparts in terms of partition function approximation under all settings. However we see that when the field meta-parameter is small ($\omega_F = 0.05$) and the interaction meta-parameter is large ($\omega_I \geq 0.8$), the convex approximations do better than Bethe in terms of marginal probabilities estimates. These results are consistent with previous studies [2, 4, 13].

Among the convex free energies optimal L2 approximation of the Bethe free energy (convexBethe-$\mu$) does better than optimal L2 approximation of the Bethe counting numbers (convexBethe-$c$) in most of the range of parameters. convexBethe-$\mu$ does not perform well in the low interaction regime (small $\omega_I$). This is presumably due to the fact that it is not forced to be variable-valid, and thus will not be exact for independent (or close to independent) models.

When averaging across all parameter settings, convexBethe-$\mu$ yields the best performance among the convex approximations. We conclude that if one seeks a convex $\mathbf{c}$ that performs across a range of parameters, it is advantageous to approximate the Bethe entropy function rather than its counting numbers. However, this comes at a price of lower approximation quality in some regimes. Regarding computation time, both heuristics require extra calculations. We note however, that as the computation does not depend on the model parameters, these extra cal-culations need to be performed only once for each model structure. Once the counting numbers are determined, the optimization of the approximate free energy is exactly the same as in standard BP. In addition, the performance of the convexBethe-$\mu$ approximation depends on the quality of the estimation of the matrix $\mathbf{A}$ in Eq. (15). This introduces a trade-off between the cost of the MCMC simulation and the quality of the approximation, which can be controlled by the MCMC convergence threshold.

### 5.2 Adaptive Approximations

The approximations we examined so far were based on the structure of the model alone, and were not tuned to its parameters. Intuitively, a good approximation should assign more weight to "stronger" interactions than to weaker ones. Indeed, Wainwright *et al.* [13], introduce a method for finding the optimal weights in TRW (denoted *TRW-opt*). The TRW entropy upper bounds the true entropy $H(\boldsymbol{\mu})$ and thus the corresponding variational approximation in Eq. (5) results in an upper-bound on the true partition function. Wainwright *et al.* thus seek counting numbers that minimize this upper bound.

Here we present a different and simpler approach for adaptively setting the counting numbers. As in the previous section, our motivation is to approximate the performance of the Bethe approximation via convex free energies. One strategy for doing so is to consider only $\mathbf{c}$ where $H_{\mathbf{c}}(\boldsymbol{\mu}) \geq H_{\mathbf{b}}(\boldsymbol{\mu})$. This implies that the optimum in Eq. (5) will always upper bound the Bethe optimum. To come as close as possible to the Bethe optimum, we can then min-imize the optimum of Eq. (5) over the counting numbers **c**.

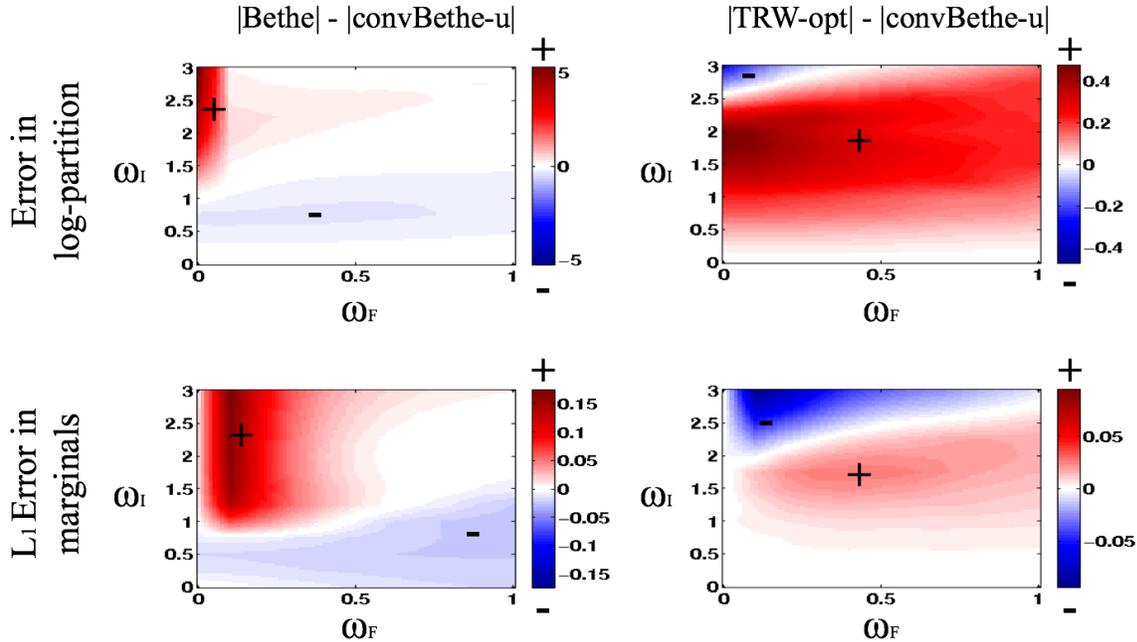

Figure 5: Comparison of partition function and marginal probabilities estimates for several free energy approximations. In each panel the $x$-axis spans values of $\omega_F$ and the $y$-axis spans values of $\omega_I$ (attractive setting). Each pixel shows the difference between average absolute errors over 20 random trials for these meta-parameters. That is, red pixels show parameterizations where the error of the first approximation is larger, and blue pixels show parameterizations where the error of the second approximation is larger.

How can we constrain **c** to upper bound the Bethe free energy? It turns out that there is a simple condition on **c** that achieves this, as we show next.

**Proposition 5.1:** *If* $\mathbf{c} \in \mathcal{C}_{vv}$ *then the difference between the approximate free energies can be written as:*

$$H_\mathbf{c}(\boldsymbol{\mu}) - H_\mathbf{b}(\boldsymbol{\mu}) = \sum_\alpha (1 - c_\alpha)\mathcal{I}_\alpha(\mu_\alpha)$$

*Where* $\mathcal{C}_{vv}$ *is the variable-valid subspace and* $\mathcal{I}_\alpha(\mu_\alpha) = \sum_{i \in \alpha} H_i(\mu_i) - H_\alpha(\mu_\alpha)$ *is the* multi-information *of the distribution* $\mu_\alpha$ *(see also [9]).*

Since $\mathcal{I}_\alpha(\mu_\alpha) \geq 0$ always holds, we get that if $c_\alpha \leq 1$ for all $\alpha$, then $H_\mathbf{c}(\boldsymbol{\mu})$ is an upper bound on the Bethe entropy. This property is not only sufficient but also necessary if we want to find counting numbers for which the bound holds regardless of $\boldsymbol{\mu}$. Note that the TRW counting numbers satisfy this constraint and thus the TRW approximation is also an upper bound on the Bethe free energy. Finally, we notice that due to the convexity constraints (Eq. (7) and Eq. (8)) the resulting entropy approximation is always non-negative (unlike Bethe).

We now show how to minimize the upper bound on the Bethe free energy. For this we generalize a result of Wainwright *et al.*[13]:

**Proposition 5.2:** *Let* $\tilde{F}_\mathbf{c}[\boldsymbol{\mu}, \boldsymbol{\theta}]$ *be free energy approximation with* $\mathbf{c} \in \mathcal{C}_{vv}$. *The subgradient of* $\max_{\boldsymbol{\mu}} \tilde{F}_\mathbf{c}[\boldsymbol{\mu}, \boldsymbol{\theta}]$ *is:*

$$\frac{\partial \max_{\boldsymbol{\mu}} \tilde{F}_\mathbf{c}[\boldsymbol{\mu}, \boldsymbol{\theta}]}{\partial c_\alpha} = -\mathcal{I}_\alpha(\mu_\alpha^*)$$

*where* $\boldsymbol{\mu}^*$ *maximizes* $\tilde{F}_\mathbf{c}[\boldsymbol{\mu}, \boldsymbol{\theta}]$.

Given this gradient, we can use a conditional gradient algorithm as in [13] to minimize $\max_{\boldsymbol{\mu}} \tilde{F}_\mathbf{c}[\boldsymbol{\mu}, \boldsymbol{\theta}]$ over the set of **c** that gives an upper bound on Bethe. Our algorithm is identical to TRW-opt except for the choice of search direction within the conditional gradient algorithm. In TRW-opt this involves finding a maximum weighted spanning tree while in our case it involves solving a LP ($\arg\min_c -c^T \mathcal{I}$ subject to the constraints). We denote the result of this optimization process by *convexBethe-u*. Empirically we find that this method is faster than the TRW iterative optimization algorithm and requires less calls to the inference procedure. More importantly, while finding the optimal counting numbers in TRW is computationally hard for non-pairwise models [14], our method is naturally applicable in the more general setting.

To evaluate the above adaptive strategy, we compare it with the convexBethe-$c$ approximation and with the Bethe approximation in the same setting we used for the static approximations (see Figure 4). In addition, Since the choice of the field meta-parameter $\omega_F$ greatly influences the relative performance of the approximation we conduct experiments to better understand its role. Instead of fixing the field meta-parameter and varying the coupling strength we plot a two-dimensional map where both meta-parameters are free to change (see Figure 5).

As we can see in Figure 4 both adaptive heuristics improve on the static ones. Moreover, our convexBethe-$u$ procedure is often more accurate than TRW-opt. Yet, both adaptive methods are still inferior to Bethe approximation for most models we tested, except for the particular regions we discuss above. More extensive comparison for different choices of parameters (Figure 5) reinforces the observation that the accuracy of the approximations differ under various model parameters. This suggests that given the model structure, there is no single "best" choice of counting numbers which is better under all parametric settings. We do see, however, that the Bethe approximation gives better or equivalent estimates of the log-partition function compared to the convex approximations (negative values in the map) except for the region with a very weak field meta-parameter ($\omega_F \approx 0$) and a strong interaction meta-parameter ($\omega_I > 1.5$). Furthermore, we see that the advantage of convex approximations over Bethe in marginals estimation is also in the region where $\omega_F$ is weak and $\omega_I$ is strong.

To examine the generality of these observations, we repeated the experiments described here for models with a structure of fully connected graph over 10 nodes (see Appendix A). These dense models differ from the sparse grid structured models, yet we get very similar results. We also conducted similar experiments for smaller and larger grids (see Appendix A) and for models with non-pairwise potentials (not shown), again with very similar results. We therefore believe that the conclusions we draw here are valid for a wide range of models.

## 6  Discussion

The study of convex free energies was originally motivated by the realization that loopy belief propagation was optimizing the non-convex Bethe free energy. It thus set out to alleviate the non-convexity problem in the Bethe optimization procedure, and indeed has resulted in elegant algorithmic message-passing solutions for convex free energies. Another interesting application of convex free-energies was for optimizing Bethe (or Kikuchi) free energies via sequences of local convex approximations [6]. Although this resulted in faster optimization, it still inherited the local-optima problem of the Bethe optimization. More recently, convex free energy variants were shown to be particularly useful in the context of model selection [12].

Despite these merits, in terms of quality of the approximation, convex free energies are still often not competitive with Bethe and in fact result in poorer performance over a wide range of parameter settings, as we also show here. This leads to the natural question, which we address in this work: what is the best convex approximation to the Bethe free energy?

As we have shown, there are several approaches to this problem, depending on whether we seek a set of counting numbers that is independent of the model parameters, or one that can be tuned adaptively. Our results show that convex Bethe approximations often work better than other schemes. For example, the counting numbers that approximate the Bethe entropy in $L2$ norm across all $\mu$ values often work better than other choices. Furthermore, our adaptive strategy for choosing the best counting numbers for a given model often works better than other methods such as TRW. Our adaptive procedures are also easily extendible to non-pairwise regions, unlike TRW which becomes intractable in these cases.

One might argue that it is more reasonable to directly approximate the true entropy $H(\boldsymbol{\mu})$ rather than the Bethe entropy. The main difficulty with this approach is that $H(\boldsymbol{\mu})$ is not generally known, and thus its approximations are typically quite loose. For example, it is not clear how to go about approximating it in L2 norm, as we do for Bethe here. The Bethe free energy, on the other hand, is tractable and as we show can be approximated in various ways. Thus, even though we lose by not approximating the true entropy, we gain by obtaining tighter approximations to the Bethe entropy, which typically provides good performance.

Another conclusion from our results is that variable-valid counting numbers usually outperform non-valid ones. One possible explanation for this fact is that they are guaranteed to give exact results for independent models. An interesting open question is what other constraints we can pose on counting numbers to enforce exactness in different scenarios, and whether we can optimize over the set of such numbers.

**Acknowledgements**

We thank Talya Meltzer, Shai Shalev-Shwartz, Raanan Fatal and Yair Weiss for helpful remarks. This research was supported in part by a grant from the Israel Science Foundation.

## A  Results for other graphs

We show results for adaptive approximations on fully connected graphs over 10 nodes in Figure 6. As can be seen, results are very similar to those of grids.

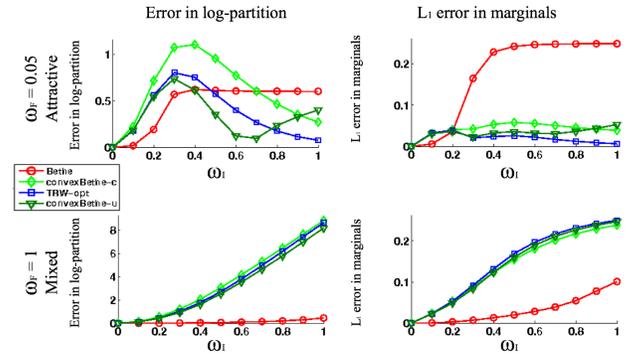

Figure 6: Comparison of estimation errors of the adaptive free energy approximations in the case of fully connected graphs. The experimental setting is similar to that of Figure 4.

Figure 7 shows similar results for $10 \times 10$ grids.

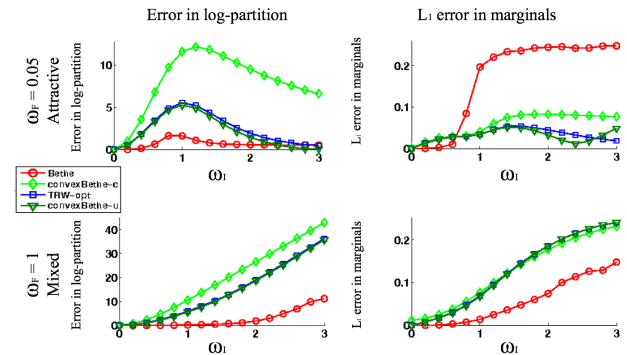

Figure 7: Comparison of estimation errors of the adaptive free energy approximations in the case of $10 \times 10$ grids. The experimental setting is similar to that of Figure 4.